\documentclass[10pt, conference, compsocconf]{IEEEtran}
\usepackage[bookmarks=false]{hyperref}
\usepackage[top=2.5cm, bottom=2.5cm, left=2.0cm, right=2.0cm]{geometry}
\usepackage{hyperref}
\usepackage{tikz}
\usepackage{paralist}
\usepackage{subfigure}
\usepackage[ruled]{algorithm2e}
\usepackage{algpseudocode}
\usepackage[cmex10]{amsmath}
\usepackage{amsfonts}
\usepackage{amsthm}
\usepackage{bm}
\usepackage{amssymb}
\usepackage{mathtools}
\usepackage{graphicx}
\usepackage{cleveref}
\usetikzlibrary{calc}
\usepackage{tikz}
\usetikzlibrary{backgrounds}
\usepackage{placeins}

\alglanguage{pseudocode}

\DeclareMathOperator*{\BigO}{\mathcal{O}}

\title{On the Performance of Network Parallel Training in Artificial Neural Networks}

\author{
    \IEEEauthorblockN{Ludvig Ericson, Rendani Mbuvha}
    \IEEEauthorblockA{School of Computer Science \& Communication \\
    KTH Royal Institute of Technology \\
    Stockholm, Sweden \\
    \texttt{\{ludv,rendani\}@kth.se}} \\
}

\date{\today}

\newcommand{\RNum}[1]{\uppercase\expandafter{\romannumeral #1\relax}}

\begin{document}

\maketitle

\begin{abstract}
Artificial Neural Networks (ANNs) have received increasing attention in recent years with applications that span a wide range of disciplines including  vital domains such as medicine, network security and autonomous transportation. However, neural network architectures are becoming increasingly complex and with an increasing need to obtain real-time results from such models, it has become pivotal to use  parallelization as a mechanism for speeding up network training and deployment. In this work we propose an implementation of Network Parallel Training through Cannon's Algorithm for matrix multiplication. We show that increasing the number of processes speeds up training until the point where process communication costs become prohibitive; this point varies by network complexity. We also show through empirical efficiency calculations that the speedup obtained is superlinear.   
\end{abstract}

\section{Introduction}

Artificial neural networks, or ANNs, are computational tools that are inspired by their biological namesake. ANNs are particularly useful when estimating functions of arbitrary complexity
using given example data\cite{Marwala:2013:EMU:2484540}. This in-turn makes them well suited for pattern recognition tasks  where they have been successfully applied in dimensionality reduction, time series prediction, and data mining. While they have long been on the fringes of the research community since their inception in the '80s, recent advances in computer and material sciences have allowed for an unprecedented growth in computation power, and thus allowing the possibility of even more sophisticated networks. One of the drawbacks of modern ANNs is the time it takes to train such a network.

As society becomes increasingly dependent on machine learning applications, real-time output from on-line training of neural network systems is critical to positive user experiences and usefulness of such systems. Thus speeding up neural network training with a parallel algorithm as we propose has become essential to the efficient deployment of machine learning applications. The case for parallelization is also reinforced by the increase in availability of multi-core CPUs and general purpose graphical processing units in recent years.

Most early work in parallelization of neural networks like \cite{651531} primarily suggest special customised hardware. This approach has certain limitations as such hardware is costly and may display suboptimal performance when utilized for other tasks. Consequently, this can be impractical for  modern scalability platforms such as HPC clusters and cloud computing infrastructure.

Dahl et al. \cite{Dahl_etal} present a Pattern Parallel Training (PPT) algorithm for parallel ANN training. In PPT, multiple ANNs are duplicated in different processes. In each process an ANN is trained on a randomly selected subset of the training examples. At the end of an epoch each process broadcasts its weight updates which are then applied to all other ANNs. This has the effect of reducing the training time by faster convergence. 

Suri et al. \cite{Suri02} use a  similar approach to \cite{Dahl_etal}, however, at the end of each full training round they  apply an evolutionary algorithm to select the best candidate networks for further training.

The approach presented here is most similar to our work is \cite{Long2008ScalableMP} who also use a Node Parallelization strategy,  while communicating the weights corresponding to ``ghost'' neurons. Their strategy requires that each process is fed with all the inputs. It therefore follows that this becomes onerous on the amount of memory required.

The contribution and novelty in our proposed approach centers on its memory efficiency in that by using Cannon's algorithm it does not require all the inputs to be fed to each process and that no specialized hardware is required. 
The remainder of this paper is organised as follows: Section \ref{mth} describes Artificial Neural Networks, Backpropagation and our proposed implementation of Network Parallel Training. Section \ref{exp} sets out our experiments. Section \ref{res} discusses the experimental results. We conclude our paper in section \ref{con} 

\section{Methods}\label{mth}
\subsection{Artificial Neural Networks}

An ANN is defined by a weighted, directed graph $G = (V, E)$, where the vertices are likened to neurons; edges to ``synapses,'' the connections between neurons. The weights are sensitivities, e.g. a highly-weighted connection will contribute more to the excitation of a neuron. We will concern ourselves only with fully-connected feed-forward networks. A feed-forward network is one in which the graph is acyclic, and so the neurons can be subdivided into layers. Being fully-connected, all layers are maximally connected to the next. The output $z_j$ of some neuron $j$ is thus a function of the output of the previous layer. We define it as $$ z_j = f_j(\mathbf{x}) = h(\sum_{\{i,j\} \in E} w_{ij} z_i + b_j), $$ where $h$ is some non-linear activation function, $g_i$ is the output of the $i$th neuron in the previous layer, $w_{ij}$ is the weight for that connection, and $b_j$ is the neuron's bias. A choice for the non-linearity function $h$ that has gained popularity in recent years is the ReLU, or \textit{rectified linear unit}. We will use the leaky ReLU, defined as $$h(x) = \begin{cases}
    x & \mbox{if } x > 0, \\
    0.01x & \mbox{otherwise}.
\end{cases}$$

The activation can be described in terms of a matrix multiplication. Let $W_i$ be the weights of $i$th layer, with each neuron as a column of weights. The activations $Z_i$ for that layer on a row-wise subset $X$ of the training data is then $$ Z_i = h(XW_i + \mathbf{b}_i). $$

\subsection{Training Neural Networks}

Training a neural network is done with \textit{stochastic gradient descent}, or SGD. In SGD, the gradient of the loss function $L$ is computed for a single sample. The network parameters are then updated using the expression $\bm\theta' = \bm\theta - \eta \nabla L$, where $0 < \eta < 1$ is the learning rate. 
This process is repeated until some convergence criterion is satisfied.

The loss function is typically the $L_2$ distance of the network output and the ground truth label, and some regularization, i.e. given ground truth label $\mathbf{y}$, network output $\mathbf{f}_{\bm\theta}$, and regularization strength $\lambda$, we have $$ L(\mathbf{x}, \mathbf{y}) = |\mathbf{y} - \mathbf{f}_{\bm\theta}(\mathbf{x})|_2 + \lambda |\bm\theta|_2,$$ where $|\cdot|_2$ is the Euclidean distance (i.e. $L_2$ distance.)

A common way to extend SGD is through \textit{mini-batch} processing, where a batch of samples are processed, and the gradient update is computed by the mean of each sample's gradient. A further improvement is momentum learning, where consecutive updates in the same direction contribute to ``momentum'' in that direction.

\subsection{Network Parallel Training}

\begin{figure}
    \centering
    \includegraphics[width=0.3\textwidth]{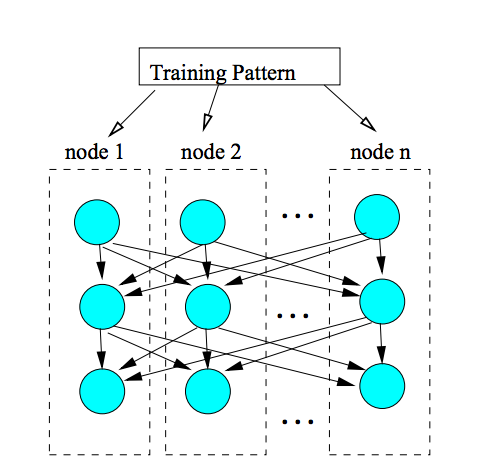}
    \caption{Illustration of the unit division in Network Parallel Training \cite{Dahl_etal}.}
    \label{fig:netsplit}
\end{figure}

Network Parallel Training (NPT) \cite{Dahl_etal} is sometimes referred to as Unit Parallelization. In NPT, one divides the network units within each layer across different processes. In its simplest form, each process is responsible for calculating activation of a single unit in each layer \cite{pethick2003parallelization}. In a more practical setup the units are divided in a linearly load balanced manner. See figure~\ref{fig:netsplit} for an example.

In our approach, the strategy is to compute the matrix-matrix multiplication $XW_i$ for each layer in parallel using Cannon's algorithm for matrix multiplication, then compute the activation output for that unit. The shifting of the local matrices within Cannon's algorithm as defined in algorithm \ref{alg:cannon} allows for the partial products required for the calculation of activations in each unit to be available within the corresponding process. 

\subsection{Cannon's Algorithm}
Cannon's algorithm is a popular parallelization strategy for matrix multiplication\cite{Cannon:1969:CCI:905686}. The algorithm efficiently computes matrix multiplications by shifting rows and columns of the candidate matrices between processes based on a defined set of rules. Algorithmm \ref{alg:cannon} shows Cannon's algorithm of two candidate matrices $A \ (M \times N)$ and $B\ ($N$ \times K)$ within a grid of $p \times q$ processes. In the context of Backpropagation, Cannon's algorithm effectively allows communication of weights between nodes that are placed within different processes. 

\begin{algorithm}
\SetAlgoLined
\DontPrintSemicolon
\KwData{Number of parts $N$, sub-matrices $A, B$}
\KwResult{Local part $C$ of global matrix multiplication}
\Begin{
    Shift rows of $A$ such that the $m$th row of $A$ shifts $m-1$ positions to the left \;
    Shift columns of $B$ such that the $n$th column of $B$ shifts $n-1$ positions to upwards \;

    \For{$i \in \{1, 2, \dots, N$\}}{
        $C \longleftarrow C + A B$ \;
        Shift rows of $A$ one process to the left, and columns of $B$ one process upwards \;
    }

    Undo initial shifts in $A$ and $B$
}
 \caption{Cannon's Matrix Multiplication}
 \label{alg:cannon}
\end{algorithm}

\subsection{Implementation}

We implement our Neural Network in C using the Message Passing Interface (MPI) \cite{Snir:1995:MCR:546703}. We use the $MPI\_Cart\_create$ function to create a virtual Cartesian grid topology that enables parallelization as depicted in \cref{fig:netsplit}.

\begin{figure}[ht]
    \centering
    \includegraphics[width=0.35\textwidth]{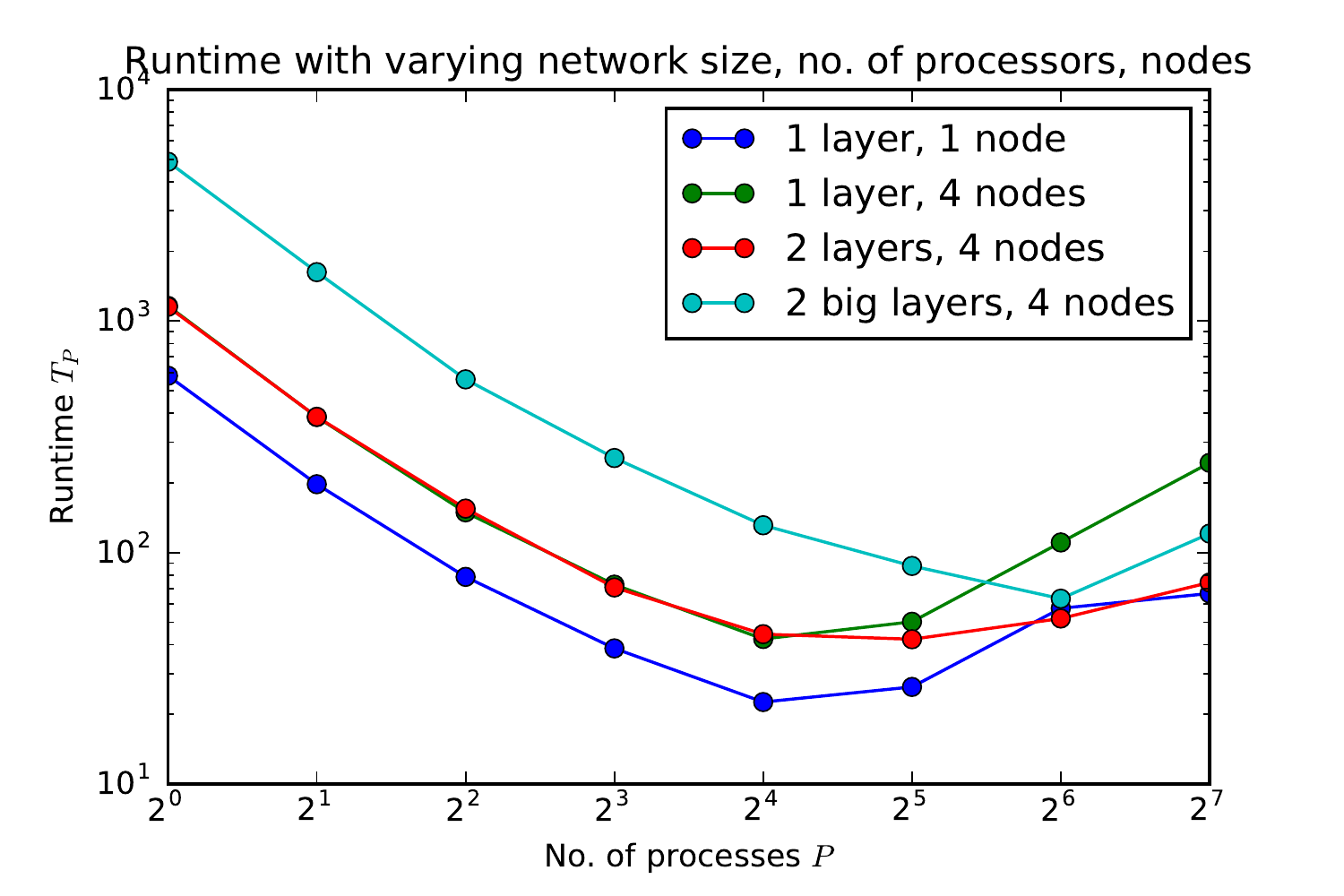}
    \caption{Plot of the runtimes $T_P.$}
    \label{fig:runtimes}
\end{figure}

\begin{figure}
  \centering
  
    \includegraphics[width=0.35\textwidth]{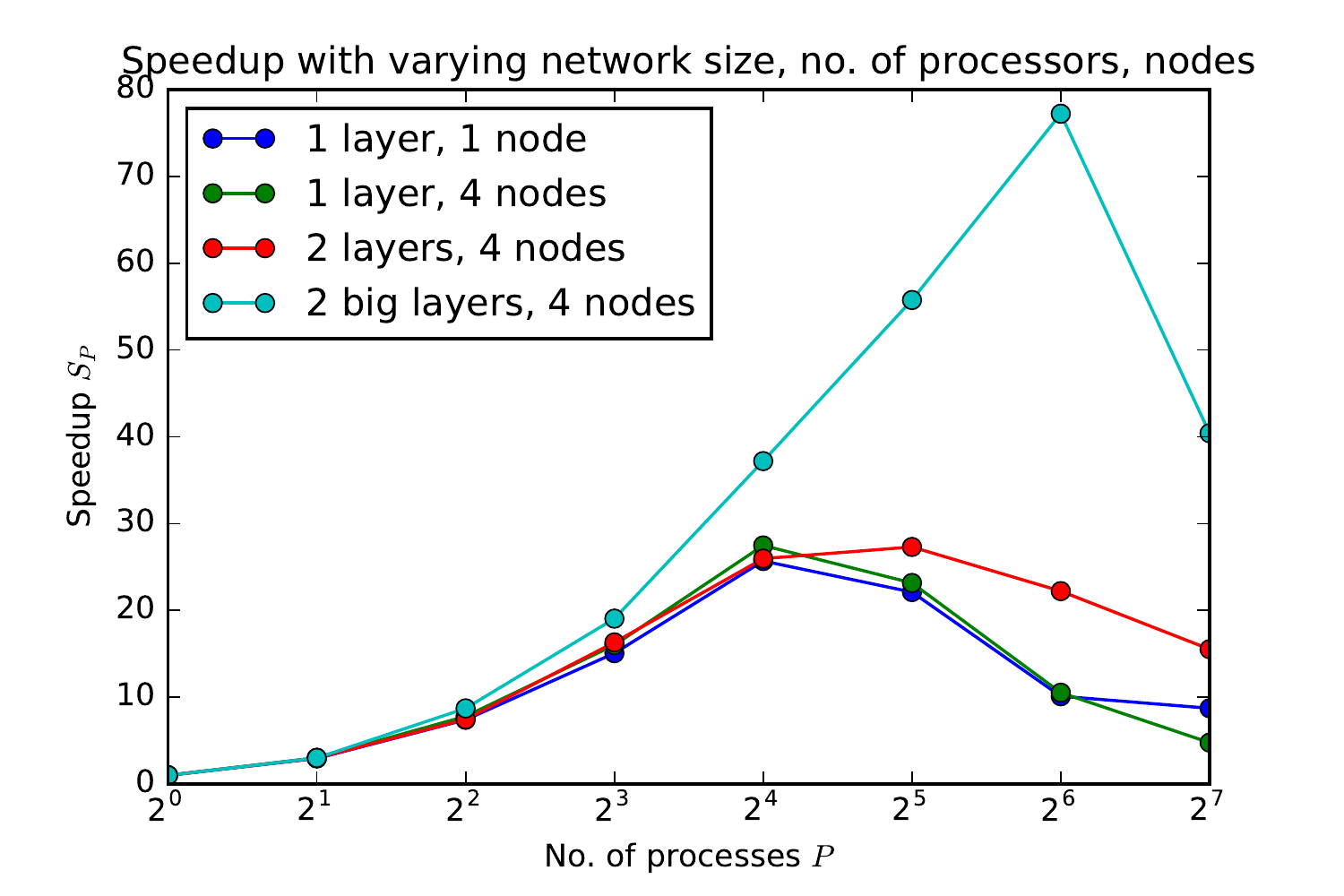}
    \caption{Plot of speedup $S_P$. We define the speedup at $P$ processes as $S_P = \frac{T^*_S}{T_P}$where $T_P$ is the execution time of our algorithm with $P$ processes. $T^*_S$ is the ``fastest known sequential version'', since we do not know of any better algorithm, we let $T^*_S = T_1$.}
    \label{fig:speedups}
  
\end{figure}  
\begin{figure}
    \centering
    \includegraphics[width=0.35\textwidth]{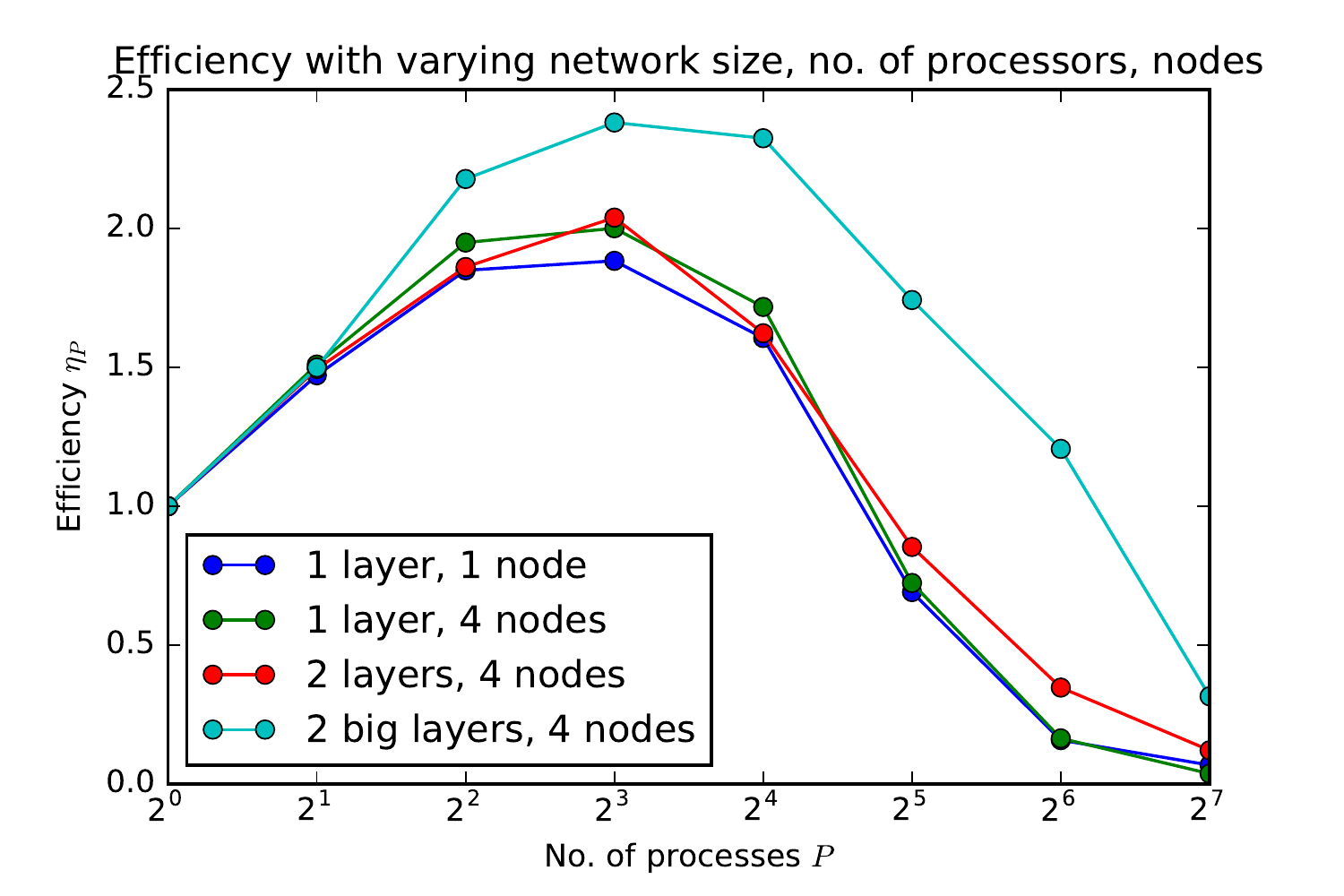}
    \caption{Plot of the efficiency $\eta_P = \frac{S_P}{P}.$
    \label{fig:efficiency}
  }
\end{figure}

\begin{table*}[ht]
    \centering
    \begin{tabular}{r|r|r|r|r}

\bf $P$ & \bf 1 layer, 1 node & \bf 2 layer, 1 node & \bf 2 layer, 4 node & \bf 2 big layers, 4 node \\ \hline
1 & 25.69 & 51.46 & 51.00 & 215.88 \\
2 & 8.73 & 17.04 & 17.07 & 71.98 \\
4 & 3.47 & 6.60 & 6.85 & 24.78 \\
8 & 1.71 & 3.21 & 3.13 & 11.33 \\
16 & 1.00 & 1.87 & 1.96 & 5.80 \\
32 & 1.16 & 2.22 & 1.87 & 3.87 \\
64 & 2.54 & 4.89 & 2.30 & 2.80 \\
128 & 2.94 & 10.80 & 3.28 & 5.34

    \end{tabular}
    \caption{Relative performance of the different workloads as measured in wall-clock time with the baseline as a  1 layer network with on single node with 16 processes.}
    \label{tab:data}
\end{table*}

\section{Experiments}\label{exp}

\subsection{Training Data}

The data was created artificially by imposing a linear relationship between the inputs $X_{ij}$  and the outputs $Y_{ik}$. The inputs are created by sampling random numbers uniformly between 0 and 200, with the bias input set to 1. The output vector is created using
\begin{align*}
    Y_{ik}=((7(k+1) \bmod 1000)+ \\( 3k \bmod 100))X_{ik}.
\end{align*}
Interestingly, since 7, 3 and 1000 are all pairwise coprimes, this forms a basis for all linear functions $y=kx+m$ with parameters $k,m \in [0, 1000)$.

\subsection{Experiment Setup}

In order to evaluate the performance of our NPT algorithm we created the following networks of different sizes and complexity:
\begin{enumerate}
    \item A One hidden layer network with the number of input and output units $q=128$
    \item A Two hidden layer network with the number of input and output units $q=128$
    \item A Two hidden layer network with the number of input and output units $q=256$ - we refer to this as the ``big'' network.
    
\end{enumerate}

We used an MPI barrier to synchronize before starting the training routine, and immediately stored the wall-clock time. The same was done after training completed. The runtime data is presented as relative runtimes in table~\ref{tab:data}.

\section{Experiment Results}\label{res}

The networks above were run on an HPC cluster with an increasing number of processors. The resulting runtimes in seconds are shown in figure \ref{fig:runtimes} together with the relative performance statistics in table~\ref{tab:data}. It can be seen that for the single layer networks the runtimes decrease as processes increase until we obtain the optimal point with 16 processes; after this point communication cost becomes prohibitive resulting in increasing runtimes. As the number of nodes are increased from 1 to 4, the runtime also increases which might imply the additional cost of redundant communication and startup related to the unused nodes. 

Increasing the number of layers $l$  from 1 to 2 increases the optimal number of processes from 16 to 32 which gives empirical evidence that the runtimes are linear in the number of layers $l$ .
The runtimes for the ``big'' network  increase quadratically relative to the two layered network on the same number of nodes 
Thus the optimal number of processes for the ``big layer'' network is 64. 

It is rather surprising that the runtimes of the two layered network are lower than those of the single layered
after 32 processes by a significant margin on the same number of nodes. This is a point that will require further investigation.

 The speedup curve in figure \ref{fig:speedups} shows similar results as in the runtime plot with exactly the same optimal process parameters. The efficiency plot in figure \ref{fig:efficiency} shows that it is more processor efficient to have 8 processors - above this the overall benefit (increase in speedup) of having additional processes decreases.
 
 An important observation from the efficiency plots is that the  processor efficiency exceeds 1 in certain regions - this implies that the seedup is superlinear \cite{Wilkinson:1998:PPT:289352}. Thus meaning - adding additional processes until a certain point actually increases the speedup possibly due to reduction in RAM access times.

\section{Conclusions and Remarks}\label{con}
An implementation of Network Parallel Training using Cannon's algorithm was proposed. Running times, Speedup and Efficiency were evaluated on networks of varying complexity. Results show that when using this strategy significant speedup is obtained. This effect is even more pronounced for more complex networks. We have also showed that there are regions where such speedup is superlinear thus increasing the efficiency of resource utilization.

It would be interesting to explore different means of parallelization other than a per-unit based split of the weight matrices. The topologies tested here were of relatively limited size. Parallelized neural networks are mostly necessary once the networks are of unusually large size, otherwise communication time is going to far outweigh any benefits of parallel execution due to the many data dependencies.

The data model used was a simple linear function depending only on one variable. While it was not relevant for the scope of this work, implementing a more sophisticated means of data generation would be an interesting next step.

Another aspect to explore is how the local matrix multiplication can be improved. Perhaps offloading the matrix multiplication to a graphics processing unit would be beneficial. 

\section*{Acknowledgements} \label{ac}
The computations in this work were performed on resources provided by the Swedish National Infrastructure for Computing (SNIC) at PDC Centre for High Performance Computing (PDC-HPC).
\FloatBarrier
\bibliographystyle{unsrt}
\bibliography{lib}

\end{document}